\pdfoutput=1

\documentclass[letterpaper, 10pt, conference]{ieeeconf}  %

\IEEEoverridecommandlockouts                              %

\usepackage{graphicx}          %
\usepackage{capt-of}           %
\usepackage{amsmath,amssymb}   %
\usepackage{booktabs}          %
\usepackage{cite}              %
\usepackage[hidelinks]{hyperref}
\usepackage{xcolor}            %
\usepackage{tcolorbox}         %
\let\labelindent\relax          %
\usepackage{enumitem}          %
\usepackage{algorithm}
\usepackage{algpseudocode}
\usepackage{placeins}

\algtext*{EndIf}
\algtext*{EndLoop}
\algtext*{EndProcedure}

\graphicspath{{figures/}}

\newcommand{\sysname}{{\sc ssf}}
\newcommand{\seg}{\ensuremath{M_{\text{\scshape seg}}}}
\newcommand{\feat}{\ensuremath{M_{\text{\scshape feat}}}}
\newcommand{\vlm}{\ensuremath{M_{\text{\scshape vlm}}}}
\newcommand{\Mthreed}{\ensuremath{M_{\text{\scshape 3d}}}}
\newcommand{\apick}{\ensuremath{a_{\text{pick}}}}
\newcommand{\aopen}{\ensuremath{a_{\text{open}}}}
\newcommand{\aobserve}{\ensuremath{a_{\text{observe}}}}
\newcommand{\anavigate}{\ensuremath{a_{\text{navigate}}}}
\newcommand{\pomdp}{{\sc pomdp}}
\newcommand{\skills}{$\mathcal{A}$}
\newcommand{\obm}{\textsc{obm}}
\newcommand{\occ}{\ensuremath{B}}

\newenvironment{tightlist}
  {\begin{itemize}[topsep=1pt, partopsep=0pt, itemsep=1pt, parsep=0pt,
      leftmargin=1.5em]}
  {\end{itemize}}

\newif\ifanonymous
\anonymousfalse

\title{\LARGE \bf
Knowing Where to Look: Interactive Object Search via Spatial-Semantic Fusion
}

\ifanonymous
\author{Anonymous Authors%
}
\else
\author{Aditya Agarwal$^{*1}$, Lucy Cai$^{*1}$, Kelly Lu$^{1}$,
        Travis Manderson$^{1}$, \\
        Nicholas Roy$^{1}$, Tom\'as Lozano-P\'erez$^{1}$,
        and Leslie Pack Kaelbling$^{1}$ \\[2pt]
        {\small Project website: \textcolor{blue!65!black}{\url{https://skymanaditya1.github.io/objsearch-website/}}}%
\thanks{$^{*}$Equal contribution.}%
\thanks{$^{1}$The authors are with the Massachusetts Institute of
        Technology, Cambridge, MA 02139, USA
        {\tt\small \{adityaag, jcai1, kellylu, travislm, nickroy, tlp,
        lpk\}@mit.edu}}%
}
\fi

\IEEEaftertitletext{%
  \begin{minipage}{\textwidth}
    \centering
    \includegraphics[width=1\textwidth]{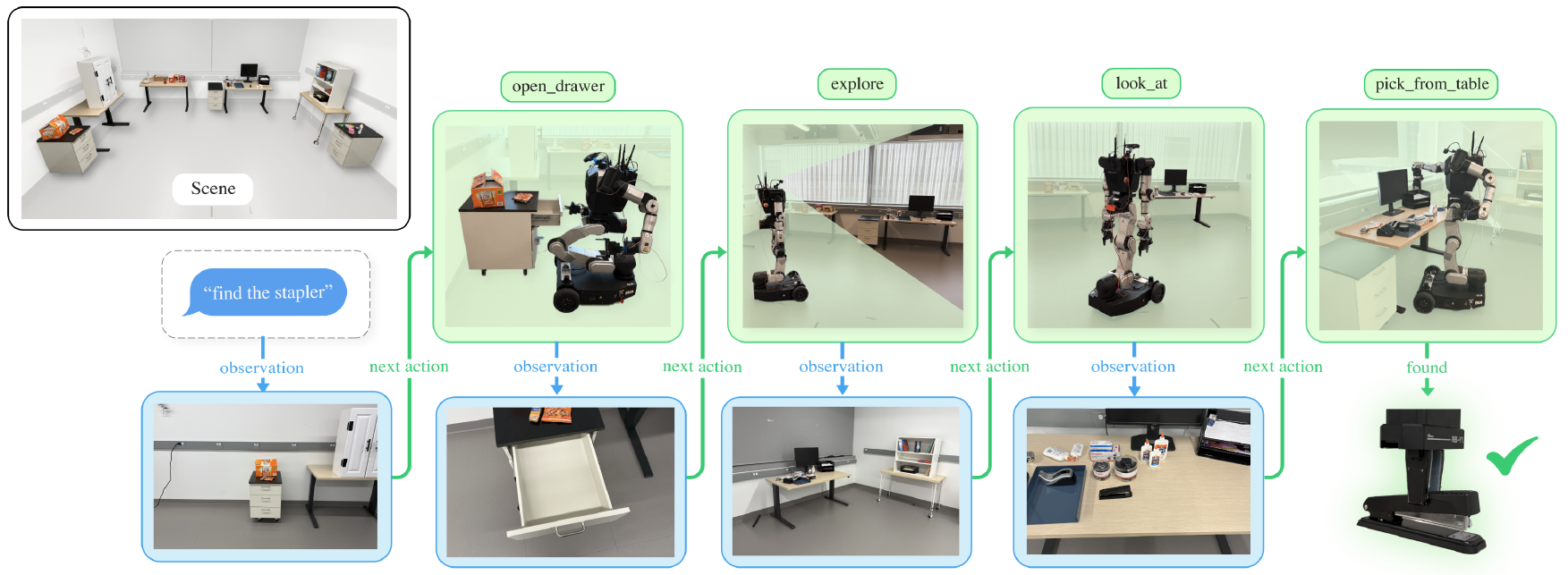}
    \vspace{-12pt}
    \captionof{figure}{\textbf{Belief-driven interactive object search.} Given
      the query \textit{``find the stapler,''} \sysname{} maintains a
      combined semantic-geometric belief field that drives action
      selection. The robot takes observations (\textit{blue}) after
      each action (\textit{green}), updating its belief until the
      target is retrieved.}
    \label{fig:teaser}
  \end{minipage}%
}

\begin{document}

\maketitle
\thispagestyle{empty}
\pagestyle{empty}

\begin{abstract}
When searching for an object, humans naturally combine semantic intuition -- reasoning about where the object is likely to be stored -- with geometric awareness of what parts of the space are visible, occluded, or unexplored. Effective robot object search requires the same joint reasoning, yet existing methods rely on only one: geometric methods maintain principled spatial beliefs but lack deep semantic understanding, while language-model based methods leverage commonsense knowledge but maintain no explicit belief over space.
We argue that jointly reasoning over semantic and geometric evidence in an explicit 3D belief enables human-like interactive object search that outperforms either modality alone.
To test this claim, we build a system that accumulates per-object VLM plausibility scores and geometric occupancy evidence via Bayesian log-odds filtering into a unified 3D voxel belief field.
High-belief regions in the combined belief field drive action selection at each step: taking a closer look from a different viewpoint, navigating toward an unexplored region of the workspace, or physically interacting with a closed cabinet.
We deploy this system end-to-end on a bimanual mobile humanoid and find that it outperforms geometry-only search and LLM-based planning methods.
\end{abstract}

\section{INTRODUCTION}

Consider a robot tasked with finding a stapler in an office supply
room, or an apple in a snack area.  The stapler might be on an open
desk, inside a closed cabinet, or tucked behind a cluttered pile of
office supplies.  The apple might be inside a snack box with
other food items, or concealed on a pantry shelf.  In both cases,
the robot cannot see everything at once: some regions are occluded
from its current viewpoint, while others are concealed within closed
containers or cabinets that must be opened to reveal their contents.
Efficiently completing these tasks requires the robot to reason about
where the target is likely to be, and decide what to do next. Doing so
may require looking from a different viewpoint, navigating to a
different region, or physically interacting with a container to reveal
its contents.

Previous approaches to object search fall into two broad paradigms, each
offering complementary strengths. One class focuses on probabilistic spatial
reasoning, explicitly modeling distributional information about the
potential locations of objects of a given
type and reasoning decision-theoretically about optimal action
strategies that include both physical and observation
actions~\cite{wandzel2019multiobject,zheng2023genmos,zheng2021cospomdp,chen2023pomdp,wong2013manipulation}. These
methods are well-founded but, with only a few
exceptions~\cite{wong2013manipulation,kollar2009utilizing}, do not incorporate any
common-sense information about where objects of various types are
likely to be found, relative to other objects.
The other class of methods leverages the rich common-sense information
available in large language and vision-language
models~\cite{honerkamp2024language,wang2025curiousbot,jiang2024roboexp} to
direct the robot's search process, generally by encoding the 3D scene
in a text-serialized representation. These methods prove excellent at
exploiting semantic information but cannot reason effectively about
spatial exploration, visibility, or reachability.

In this work, we combine the strengths of these two approaches, into
\sysname{}, which maintains a spatial belief representation that fuses
common-sense object location information from a vision language model
with geometric evidence from occupancy raycasting into a unified 3D
voxel belief field, and selects actions decision-theoretically based
on this representation. We implement and demonstrate the full system
on a bimanual wheeled humanoid in a real-world office environment.

Concretely, our contributions are: (1) a method to explicitly extract
semantic object-type occurrence information from a VLM and convert it
into a grid-based belief representation; (2) a recursive Bayesian
belief-update definition that fuses information from 3D spatial
sensing and the semantic information grid; and (3) a complete
implementation on a wheeled humanoid robot that actively searches for
objects satisfying an open-ended natural-language description, using
an information-theoretic criterion to govern its search,
combining learned skills for opening containers with pre-existing
general-purpose methods for navigation and pick-and-place.

Through real-world experiments on a bimanual humanoid robot in
room-scale settings, we find that \sysname{}
outperforms representative geometry-based and foundation-model-based
baselines in tasks at multiple difficulty levels in problems requiring
the retrieval of office objects and snacks.

\begin{figure*}[t]
    \centering
    \includegraphics[width=1\textwidth]{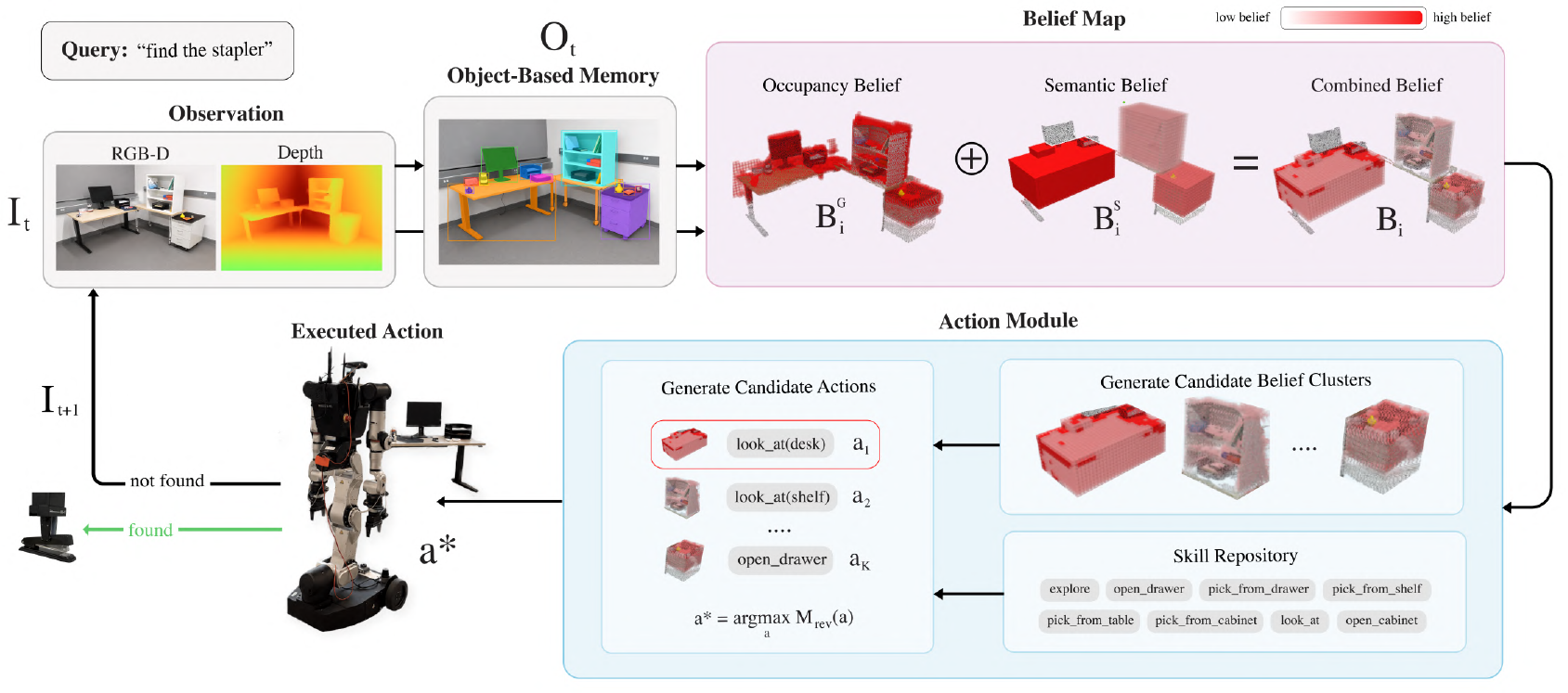}
    \vspace{-12pt}
    \caption{\textbf{System overview.} Given a natural-language query
      $q$ (e.g., \textit{``find the stapler''}) and an RGB-D
      observation $I_t$, \sysname{} maintains two persistent
      structures: the object-based memory $O$, which tracks object
      hypotheses and their per-object semantic log-odds, and the
      \textbf{Belief Map} $B$, which fuses geometric evidence $G$ from
      occupancy raycasting with semantic evidence $S$ derived from the
      \obm{}'s VLM scores. 
      \textbf{Action Module} spatially clusters high-belief cells in
        $B$, generates one candidate action per selected cluster, and
        selects the action $a^*$ with the highest expected revealed belief mass \(M_{\mathrm{rev}}\).
        The corresponding skill is dispatched from the
        \textbf{Skill Repository} (here, \texttt{look\_at}).
        The observation $I_{t+1}$ returned by the action updates $O$
        and then $B$.}
    \label{fig:architecture}
    \vspace{-12pt}
\end{figure*}

\section{RELATED WORK}
\label{sec:related_work}
The problem of searching for objects in a relatively unknown space
with a mobile robot has a long history. Methods differ in terms of the
robot's capabilities, the types of information used to inform the
search, and the framing and implementation of the search
strategy.

The majority of object-search robots are simply able to move through
space, directing a fixed or steerable camera on a mobile base. More
capable systems~\cite{wong2013manipulation, bejjani2021occlusion, wang2025curiousbot} {\em interact} with the environment, generally to
remove occluding objects or open containers, to reveal previously
unobserved space.

Object search is fundamentally a partially observable problem, which
means that any effective strategy will have to have some form of
internal state or memory. Most simply, it can be a recurrent
policy~\cite{ye2021auxiliary, wijmans2020ddppo} although such
policies can be difficult to train. It is generally more effective to
have an explicit memory, which can be of many types, including a
simple 2D map of locations that have been observed so
far~\cite{chaplot2020semexp, yamauchi1997frontier}; textual
descriptions of previous observations~\cite{honerkamp2024language,
  wang2025curiousbot}; or 3D oct-tree representations with
probabilistic semantics~\cite{zheng2023genmos, zheng2021cospomdp}.

The choice of memory representation is usually closely tied to the
source of information used by the system. These sources may include voxel
occupancy information derived from 3D scans~\cite{moravec1985high,
  thrun2005probabilistic}, simple object-type co-occurrence
statistics~\cite{wong2013manipulation, zheng2022towards}, and more
general common-sense semantic information from feature
representations~\cite{gadre2022cows, gu2024conceptgraphs},
LLMs~\cite{zhou2023esc, shah2023navigation}, and
VLMs~\cite{yokoyama2024vlfm, majumdar2022zson}.

Finally, methods can be differentiated based on their action-selection
strategy. A few approaches embed the object-search problem in a larger
\pomdp{} formulation in which the robot has a higher-level objective that
may demand locating objects as subgoals; these formulations are
typically solved using a general-purpose belief-space planning
strategy that can enable recursive subproblem solving (e.g., finding a
tool to open a container to find another object) but they are quite
computationally difficult and have only been demonstrated in simple
simulations~\cite{wong2013manipulation}. More typical are systems
that are limited to addressing a single object-search problem at a
time. These approaches range from specifically trained neural-network
policies~\cite{ramrakhya2022poni, yang2018visual} to methods that
reason about the information gain of observation actions and greedily
pursue the next best view until the object is
located~\cite{aydemir2013active, atanasov2013nonmyopic}.

Our method, \sysname{}, takes explicit manipulation actions to reveal
new space, represents its observation history in terms of a 3D belief
over the target object's potential locations, fuses spatial occupancy
information with common-sense semantic information derived from a VLM, and
uses an information-theoretic criterion to determine the next best
region of space to view.

Three lines of prior work are most closely related to \sysname{}. Each incorporates some combination of probabilistic 3D belief, foundation-model semantics, and physical interaction, but none combines all three. GenMOS~\cite{zheng2023genmos} maintains a probabilistic 3D
belief and plans next-best viewpoints over it, but incorporates no
semantic information and cannot manipulate the environment, so it
must resolve all uncertainty by repositioning alone.
IPPON~\cite{qu2024ippon} adds foundation-model semantics to such a belief
and is closest to \sysname{} in formulation: it fuses semantic scores
into a 3D voxel belief via Bayesian updates and selects actions by
expected gain over that belief. It addresses object-goal {\em
navigation} only, not manipulation, however: it succeeds by finding a viewpoint from which
the target is visible, so its belief is defined only over space that
some viewpoint can resolve. The interior of a closed container is
not visible from any viewpoint; in IPPON's formulation those voxels have zero belief, so the target can neither be represented there nor
searched for there. \sysname{} assigns positive belief,
which is what allows the belief field to drive manipulation actions
such as opening a cabinet. In addition, IPPON's semantic modeling is driven by an LLM's interpretation of category names, whereas SSF leverages a VLM's ability to discriminate among different objects of the same category (e.g., different kitchen cupboards based on nearby visible objects). Finally, CuriousBot~\cite{wang2025curiousbot} and
RoboEXP~\cite{jiang2024roboexp} combine physical interaction with
foundation-model semantics, but represent the environment as a
deterministic 3D relational or action-conditioned scene graph rather
than a probabilistic belief over the target's location, and pursue
open-ended scene exploration rather than goal-directed search.

\section{PROBLEM FORMULATION AND APPROACH}
\label{sec:problem_statement}

We address a problem in which a mobile robot is given an arbitrary natural-language description, $q$, and is intended to find and pick up an object satisfying that description. The robot must have at least one manipulator, RGB camera, and (optionally) a depth sensor. It need not know anything in advance about its environment, but is given a 3D bounding box $R$ of the region (relative to its pose) that it should search.

We assume a set of available modules:
\begin{tightlist}
\item A library \skills{} of medium-level robot ``skills'' in the form
  of policies that can plan and control base motion, pick and place
  objects, and open drawers and cabinets. These policies may involve
  calls to other perception modules, e.g., for segmentation and shape
  completion.
  \item An object segmentation model \seg{} that, given an RGB image, can
    provide a set of detected objects with masks.
  \item A semantic-feature model \feat{} that, given an image region can
    provide a latent-space semantic feature vector.
  \item A vision-language model \vlm{} that, given a bounding-box annotated RGB image, can
    provide predictions about the presence of a goal object satisfying
    $q$.
  \item If no depth sensor is available, a model \Mthreed{} for
    recovering calibrated depth from RGB (a single image, or stereo
    pair, or sequence).
  \end{tightlist}

The objective is to locate an object satisfying $q$
as efficiently as possible, or to terminate with an assertion that no
such object is available. The method is ``zero-shot,'' in the sense
that, although the interaction skills and other perception models will
generally be pre-trained, there is no actual training on the search
task.

We will formulate this problem as a \pomdp{} with a belief state
representing the likelihood of an object satisfying $q$ existing at
various locations, with actions for observing from a specified
viewpoint, for removing occluding objects, and for retrieving the
target object, with observations gained through the cameras. The complexity of the state and observation space make both exact belief inference approaches and exact solutions to the optimal \pomdp{} policy computationally intractable. We focus on developing a
rich but approximate belief-state representation and a Bayesian
filtering method for belief updating based on both geometric and semantic
information. We use a greedy action-selection mechanism relative to
that belief until a goal object has been observed and retrieved.

Fig.~\ref{fig:architecture} gives an overview of \sysname{}.
In the following subsections, we describe our approaches to belief
representation and update, and to action selection.

\subsection{Belief representation and update}
\label{sec:belief_representation}

The belief state in \sysname{} consists of two components: an object-based
memory (\obm{}) $O$ of the objects it has encountered so far and a 3D spatial
representation, \occ{}, that encodes beliefs about occupancy.

\paragraph{The object-based memory} has the form of a classic data
association filter~\cite{daf} applied to 3D object
representations~\cite{obj_based_slam}, and combining spatial and
semantic information to make fusion decisions, similar
to~\cite{gu2024conceptgraphs}.  For each object hypothesis, it maintains
a representation consisting of: a partial point cloud, 3D bounding
box, a latent semantic feature vector, images and masks for each
detection, and a text label and caption.

Importantly, the \obm{} is used both to check whether we have located a goal object as well as to represent larger objects (e.g., cupboards or cabinets) that might contain the target object. The \obm{} supports unconstrained natural-language queries, returning an object in the memory if there is one that satisfies the query, as judged by \vlm{}.

Given a new observation RGBD image $I$, we use $\seg{}$ to locate a set
of detections $D$ with image masks, and use $\feat$ and $\vlm$ to
compute a latent semantic feature vector and text caption for
efficient retrieval.  Given detections $D$, a matching algorithm
combines a 3D bounding-box overlap score with semantic feature
similarity to find the best assignment of new detections to previous
hypotheses in $O$. The matched hypotheses are updated based on the new
detections, and new hypotheses are created for unmatched detections
and added to $O$.

\paragraph{The belief map} is a 3D voxel-based representation of
the likelihood of finding an object at each location, where
$\occ{}_i$ encodes the probability of the event $C_i$, that there is an object satisfying
description $q$ overlapping cell $i$. These are separate, per-cell
Bernoulli beliefs (with no constraint that they sum to 1), not a
distribution over the location of a {\em particular} object.
It is the combination of two sources of information, geometric ($G$) and
semantic ($S$), which we treat as conditionally
independent given $C_i$, the event that cell $i$ overlaps an object satisfying $q$. For simplicity of calculation,
following the standard in occupancy grid
representations~\cite{elfes1989}, we represent the {\em log odds},
$\ell = \log (p / (1-p))$ rather than the probability, so that

\begin{equation*}
\resizebox{0.95\columnwidth}{!}{$
\begin{aligned}
\occ{}_i
&= \log \frac{P(C_i \mid G,S)}{1-P(C_i \mid G,S)} \\
&= \log \frac{P(C_i \mid G)}{1-P(C_i \mid G)}
 + \log \frac{P(C_i \mid S)}{1-P(C_i \mid S)}
 - \log \frac{P(C_i)}{1-P(C_i)}
\end{aligned}
$}
\label{eq:belief}
\end{equation*}

In the case of a uniform prior on $C,$ this reduces to the sum of geometric and semantic voxel log-odds \cite{thrun2005probabilistic}. 

The geometric data is stored in a persistent occupancy grid, $G$,
which stores log-odds values, with $G_i$ encoding the probability that that cell is occupied. Given a new depth map from RGBD image I, cells that rays traverse have their occupancy likelihood decreased, cells that rays terminate in have their occupancy likelihood increased, and others maintain their previous values. The conditional $P(C_i \mid G_i)$ is $\approx 0$ when the cell is free, and a fixed constant otherwise. In practice, similar to existing works \cite{zheng2023genmos, qu2024ippon}, we assign each cell a geometric log-odds value according to its observation state: $-6.91$ for cells observed to be free, $-4.00$ for cells observed to be occupied (by an object other than one satisfying $q$), $-2.20$ for unexplored cells, and $-0.85$ for cells that were occluded in some observation, and have not yet been observed. 

The semantic data is derived from the objects in \obm{}. We maintain,
for each object $o$, a log-odds estimate, $\ell_o$, that the object contains or
conceals an object satisfying $q$. Whenever an image observation is
associated with object $o$ in the filter update, the \vlm{} is queried
with that image to obtain an estimate $p_o$ of the probability that
$o$ contains, conceals, or is an object satisfying $q$.
If the VLM scoring were
highly accurate, we could treat $p_o$ directly as an estimate of the
probability that object $o$ contains an object satisfying $q$.
However, these outputs are unreliable \cite{xiong2024can, groot2024overconfidence, gligoric2025can}, so we instead treat them as
observations to be aggregated over time. We construct a Bayes filter,
initializing $\ell_o$ to $0$ (an uninformed prior) and updating it
after each new semantic observation:
\[\ell_o := \ell_o + \log \frac{p_o}{1 - p_o}.\;\;\]

In practice, we clamp $p_o$ to $[\epsilon, 1-\epsilon]$ with
$\epsilon = 10^{-3}$, so the update remains well-defined and no
single VLM output can permanently eliminate or confirm a hypothesis. 
This update assumes that new observations of an object yield information about whether the object $o$ contains or conceals an object satisfying $q$ is conditionally independent of previous views. 
Since nearby viewpoints produce highly
correlated observations, this assumption only holds approximately. To mitigate the effects of correlated observations, we only incorporate
observations from viewpoints that are substantially different from
previous ones~\cite{javier_velez}: separated by at least 0.15\,m in
base position, $15^\circ$ in base yaw, or $10^\circ$ in head tilt,
thresholds chosen to yield empirically distinct observations.

For each cell $i$ in the grid, we let the semantic log-odds
evidence be
\[\ell(C_i \mid S) = \max_{\{o\in\obm{} \mid \text{$o$ overlaps cell
      $i$}\}} \ell_o
  \]
with a default value of $0$ for cells that do not overlap any objects.

\subsection{Action selection}
\label{sec:action_selection}

\begin{algorithm}
\caption{Interactive object search with \sysname{}. The query $q$ is a
  natural-language object description and $R$ is the region of space to
  search.}
\label{mainalg}
\small
\begin{algorithmic}[1]
\Procedure{ObjectSearch}{$q$, $R$}
  \State $O \gets$ object-based memory
  \State $B \gets$ belief grid over region $R$
  \Loop
    \State \textit{target} $\gets O.\textsc{contains}(q)$
    \If{\textit{target} $\ne$ \textsc{None}}
      \State execute $\apick(\textit{target}, B)$
      \State \Return \textsc{success}
    \EndIf
    \State \textit{clusters} $\gets B.\textsc{clusterHighBelief}()$
    \State $\mathcal{C} \gets \textsc{candidateActions}(\textit{clusters}, O)$
    \If{$\mathcal{C} = \emptyset$}
      \State \Return \textsc{fail}
    \EndIf
    \State $a^* \gets \arg\max_{a \in \mathcal{C}} \sum_{i \in F(a)} \sigma(\occ{}_i)$
    \State $i \gets$ execute $a^*$
    \State $O.\textsc{update}(i)$
    \State $B.\textsc{update}(i)$
  \EndLoop
\EndProcedure
\end{algorithmic}
\end{algorithm}

The overall operation of \sysname{} is described in
Algorithm~\ref{mainalg}. It begins by testing whether there is an
object already in the \obm{} that satisfies $q$ and, if so, selects
the \apick{} action to pick up that object, and terminates.

Otherwise, action selection proceeds in three stages. First,
high-belief cells in $B$ are clustered spatially into connected
components. 
We choose k high-belief clusters, and generate one candidate action for each identified high-belief cluster: an \aobserve{} action viewing the cluster from a new viewpoint, or, if the cluster lies inside a closed object in $O$, an \aopen{} action on that object, and select the action with the highest expected revealed belief mass across all candidates. 
Each candidate action \(a\) has an associated
footprint \(\mathcal F(a)\): the set of not yet observed cells it would
reveal—the cone of space an \(a_{observe}\) would see, or the
interior an \(a_{open}\) would expose—expressed as a set of voxel
indices.
\[
\mathcal{F}(a)
=
\left\{
i \in \mathcal{U}_t
\;\middle|\;
\widehat{\operatorname{vis}}_t(i,a) = 1
\right\}.
\]
$\widehat{\operatorname{vis}}_t(i,a)$ is a binary indicator that equals $1$ if a voxel $i$ is predicted to be visible after executing action $a$, and $0$ otherwise, and $\mathcal{U}_t$ denotes the set of voxel indices that have not yet been observed at time $t$. Third, the module selects
\[a^* = \arg\max_{a} M_{\mathrm{rev}}(a), \quad
  M_{\mathrm{rev}}(a) = \sum_{i \in \mathcal{F}_a} \sigma(\occ{}_i),\]
where $\sigma$ converts log-odds back to probability, and \(M_{\mathrm{rev}}(a)\) is the expected belief mass that action \(a\) would reveal.
If no candidate remains whose footprint could still contain the target, the search terminates with a failure. Executing the selected action returns an image $i$, which is used to update the \obm{} and then the belief grid.

Each of the actions is itself a sophisticated policy that includes
collision-free navigation and manipulation with respect to the
obstacles in $B$ and object shape information in $O$. We use greedy action selection as a practical heuristic, choosing at each step the action with the highest expected belief mass $M_{\mathrm{rev}}$.

\section{SYSTEM IMPLEMENTATION}
We have implemented an instance of \sysname{} using currently
available modules. Importantly, the modular
design of \sysname{} means that as advances are made in foundation
models and robot control policies, new implementations can be
substituted in directly, to gain overall improved performance.

Our implementation uses a Rainbow Robotics RB-Y1 bimanual wheeled humanoid, with parallel-jaw grippers. A base-mounted Livox LiDAR provides odometry, a head-mounted ZED 2i stereo camera is actuated in pan and tilt, and wrist-mounted monocular ZED cameras are used in training neural policies. Each arm supports the learned cabinet- and drawer-opening policies.

\subsection{Perception modules}
We use {\sc clip ViT-H/14}~\cite{clipradford2021learning} as \feat{}, {\sc sam3}~\cite{carion2025sam} as
\seg{}, and Gemini 3.1 Pro~\cite{gemini} for \vlm{}. We have 3D sensors so we do
not require an implementation of \Mthreed{}.

To evaluate whether there is an object in the \obm{} that satisfies
$q$, for efficiency, we begin by finding the $k$ objects $o$ whose
{\sc clip} embeddings are closest in cosine distance to the {\sc clip}
embedding of $q$. Given this set of plausible candidates we run the
more expensive Gemini query on a stored image crop with a prompt
asking whether this object satisfies $q$.

Given a new observation image $i$, we extract object detections using
the following pipeline: (1) we use a standard label catalog to query
Gemini to get candidate bounding boxes, then (2) use SAM in box-prompt
mode to get pixel-level segmentation masks. For data-association in
the \obm{} we use a weighted combination of cosine distance of {\sc
  clip} feature vectors and 3D bounding-box overlap.

Since the semantic belief depends on VLM outputs, we measured their
repeatability: querying Gemini on 10 frames (5 per environment) 10
times each, we observed that 54\% of per-object scores were identical across runs and
91\% varied by at most 0.1, with a near-deterministic ranking of
objects and the median latency was 10\,s per scoring call.

\subsection{Robot action policies}
\begin{figure}[t]
    \centering
    \includegraphics[width=1\columnwidth]{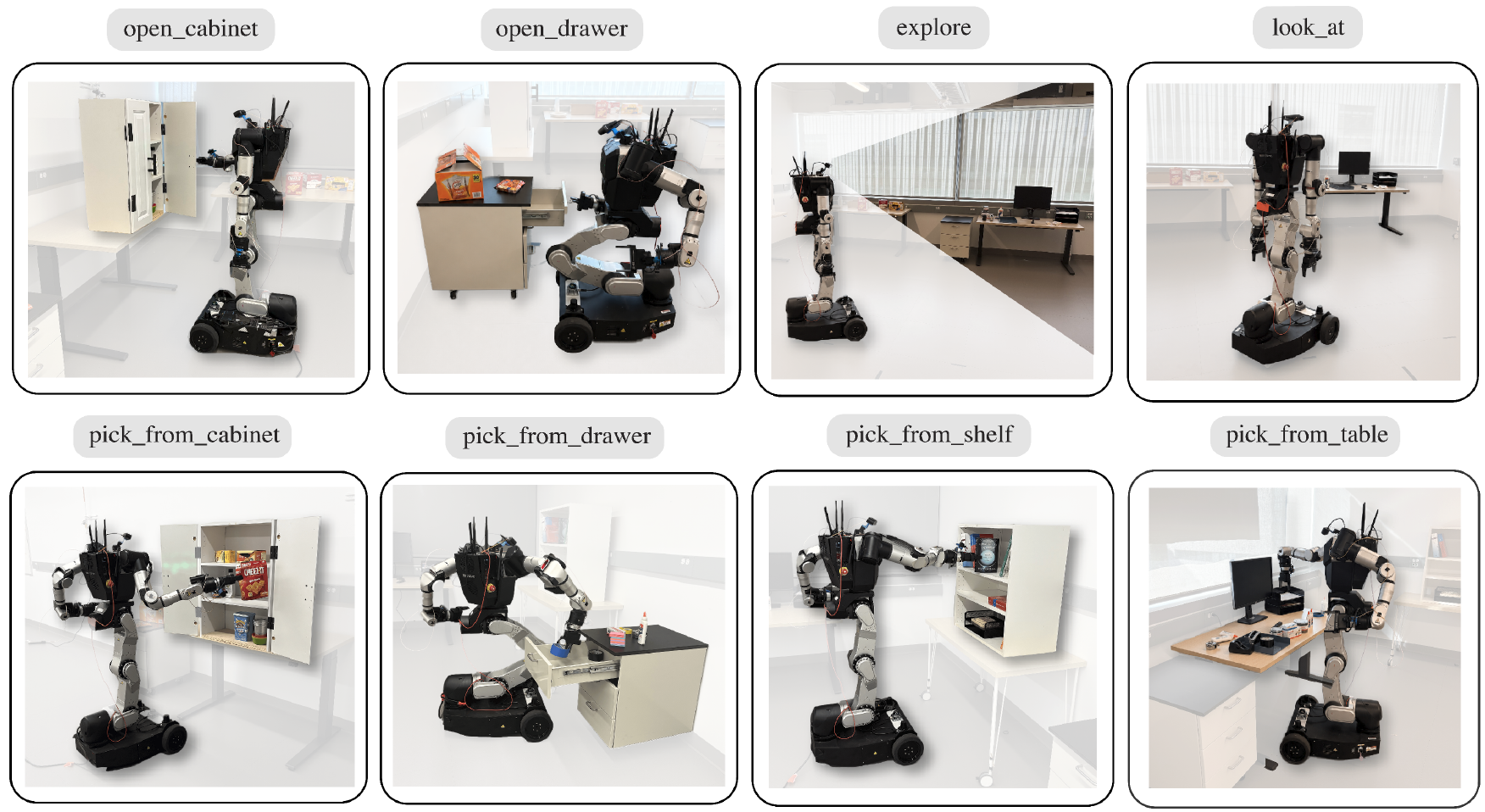}
    \vspace{-8pt}
    \caption{
\textbf{Skill library.} \texttt{open\_cabinet} and \texttt{open\_drawer} are learned via behavior cloning; \texttt{explore} navigates to frontier viewpoints; \texttt{look\_at} repositions the base, torso, and head to uncover occluded voxels. The four \texttt{pick\_*} skills (bottom row) retrieve objects from cabinets, drawers, shelves, and tables.}
    \label{fig:skills}
    \vspace{-12pt}
\end{figure}

The robot's actions are implemented using a collection of
goal-conditioned policies with substantially different implementation
strategies. They are illustrated in Fig.~\ref{fig:skills}.

The $\anavigate{}(x, y, \theta)$ action plans a path to move the
robot's non-holonomic base to a specified pose using A* on a 2D
occupancy map obtained by down-projecting the 3D voxel grid, and then
executes it with a simple feedback controller. It is never called at
the top level of the system, but is used as a component in all of the
other actions, to move the base into place.
The $\aobserve{}(v)$ action makes an inverse kinematics call to find a
base pose that will enable putting the camera at the viewpoint $v$,
calls \anavigate{} to move there, moves the head appropriately, then
captures and returns an image observation.
The $\apick{}(o)$ action makes an inverse kinematics call to find a
base pose that will enable manipulating object $o$, calls \anavigate{}
to move there, and then calls an adapted version of
TipTop~\cite{shen2026tiptop}, an open-vocabulary manipulation system
originally developed for fixed-base tabletop manipulators.
We adapt it to the RB-Y1 by parameterizing the torso and the 7-DOF
right arm for executing collision-free picking.
The $\aopen{}(o)$ action is a collection of methods, tailored
to specific container objects, that are
learned via behavior cloning using Diffusion
Policy~\cite{chi2023diffusionpolicy}.

\section{EXPERIMENTAL RESULTS}
\label{sec:result}
We design our experiments to evaluate three aspects of the system: 1) how the combined semantic-occupancy belief drives search and interaction through the workspace, 2) how the combined belief compares against occupancy-only and LLM-based baselines over two object categories and four placement settings, and 3) whether our full system integration, learned interaction skills, open-vocabulary pick-and-place, navigation, and whole-body viewpoint selection, enables a bimanual wheeled humanoid to perform interactive object search in the real world.

\begin{figure}[t]
    \centering
    \includegraphics[width=1\columnwidth]{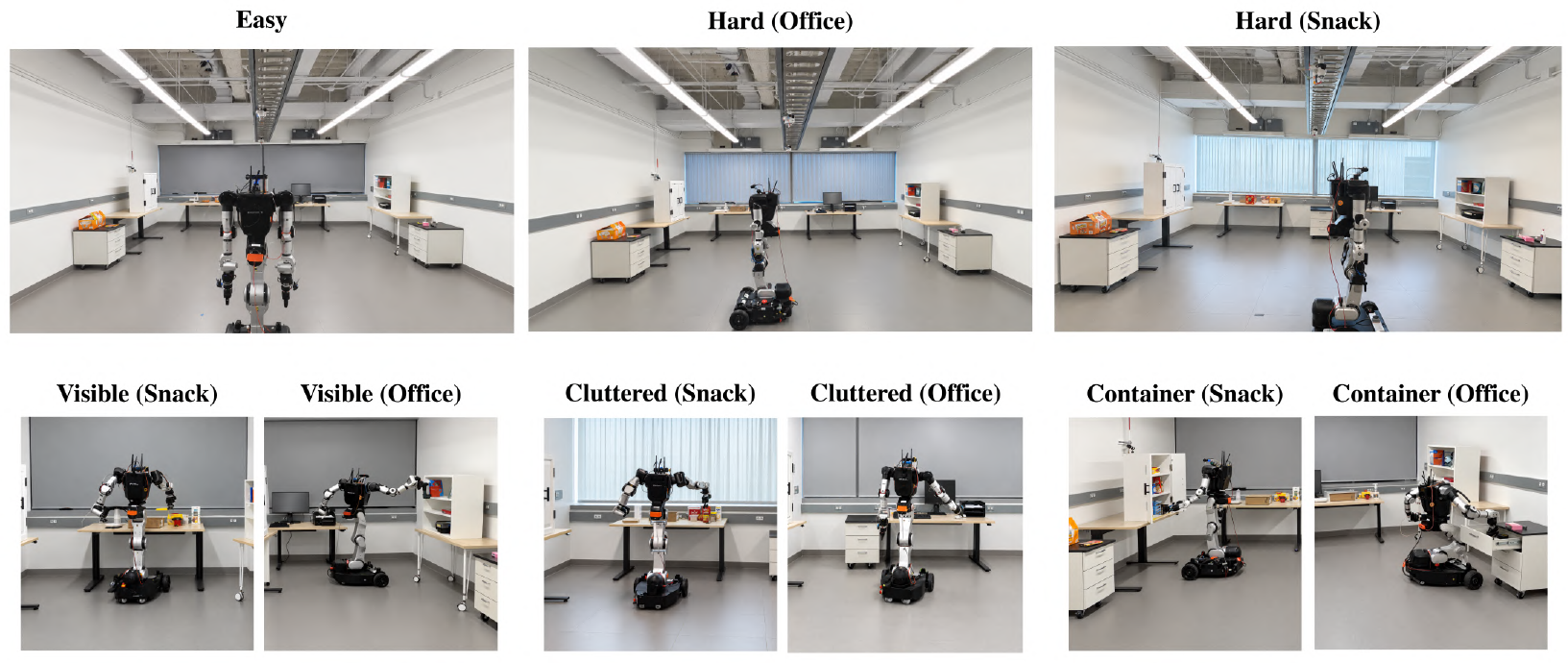}
    \vspace{-12pt}
    \caption{\textbf{Experimental setup and task settings.}
\textbf{Top:} Easy and hard initial viewpoints. The easy
viewpoint exposes most of the workspace; hard viewpoints
initially face away from the target object's office or snack area.
\textbf{Bottom:} Examples of visible, cluttered, and container
tasks in both areas.}
\label{fig:task_settings}
    \vspace{-12pt}
\end{figure}

\subsection{Real-World Setup}

We run all experiments in a real-world office of approximately 87\,m$^3$ (6.04\,m $\times$ 4.85\,m $\times$ 2.97\,m), divided into a kitchen pantry area and an office supply area. The kitchen contains a snack table (Cheez-Its, a fruit basket, crackers), a pantry cabinet of breakfast items (peanut butter, jam, tea), and a storage drawer with a snack box on top. The office contains a desk cluttered with supplies (glue, tape, staplers, binder clips, paper), a bookshelf with books and folders, and a storage drawer holding duct tape with supplies on top. An empty storage drawer separates the two areas.

\paragraph{Task Settings} We evaluate four placement settings per
object category, each testing a different aspect of search.
(1) \textit{visible}: the target is in plain sight (a book on the bookshelf; a Cheez-It box on the snack desk), requiring only retrieval.
(2) \textit{clutter}: the target is occluded within a pile of similar objects (a stapler among office supplies; a banana inside a fruit basket), requiring a closer look and visual disambiguation.
(3) \textit{container}: the target is inside a closed container (duct tape in the storage drawer; a peanut butter jar in the pantry cabinet); the robot must identify the correct container and open it.
(4) \textit{floor}: the target has fallen near a semantically related surface rather than into standard storage (a glue bottle from the office desk; a pear from the snack table). The system must place semantic mass on that surface so the object is incidentally observed when the robot orients toward it. We show a subset of the tasks and viewpoints in Fig.~\ref{fig:task_settings}.
We evaluate each setting under two robot viewpoint initializations.
This gives a total of $\mathbf{48}$ real-robot experimental trials.

\subsection{Baselines}
We compare against two adapted baselines that use different
search representations and decision rules while sharing
perception, navigation, and the skill library with \sysname{}.

\textbf{Geometry-only.} We adapt GenMOS~\cite{zheng2023genmos}
to use our full action set and replace its POUCT planner with
greedy action selection. We retain its multiplicative voxel
belief update, initialized by occupancy state: $2.0$ for
occluded, $0.3$ for unexplored, and $1.0$ for occupied cells.
Raycasting updates occupancy, while VLM detections for target object multiply the corresponding voxel beliefs by $10^3$;
otherwise, visible voxels decay by $10^{-3}$.
Retrieval is triggered when any voxel belief exceeds $10^6$ and the search terminates otherwise when the time budget is exhausted. 

\textbf{LLM-planner.} We follow the implementation in CuriousBot~\cite{wang2025curiousbot}. This baseline maintains
a semantic scene graph of detected objects and VLM-estimated
spatial relations from RGB-D observations. Given the graph,
search query, and action history, the LLM selects an object
to investigate, an \aobserve{} action to explore unseen space,
or an \apick{} action for a visible, VLM-verified target.
We add exploration to support initial views without plausible
target locations. We follow VLFM~\cite{yokoyama2024vlfm} to annotate frontiers with visible scene-graph objects and use an LLM to select the exploration frontier by matching their relevance to the query. Search terminates upon retrieval
or exhaustion of the step or time budget.

\begin{figure}[b]
    \centering
    \includegraphics[width=1\columnwidth]{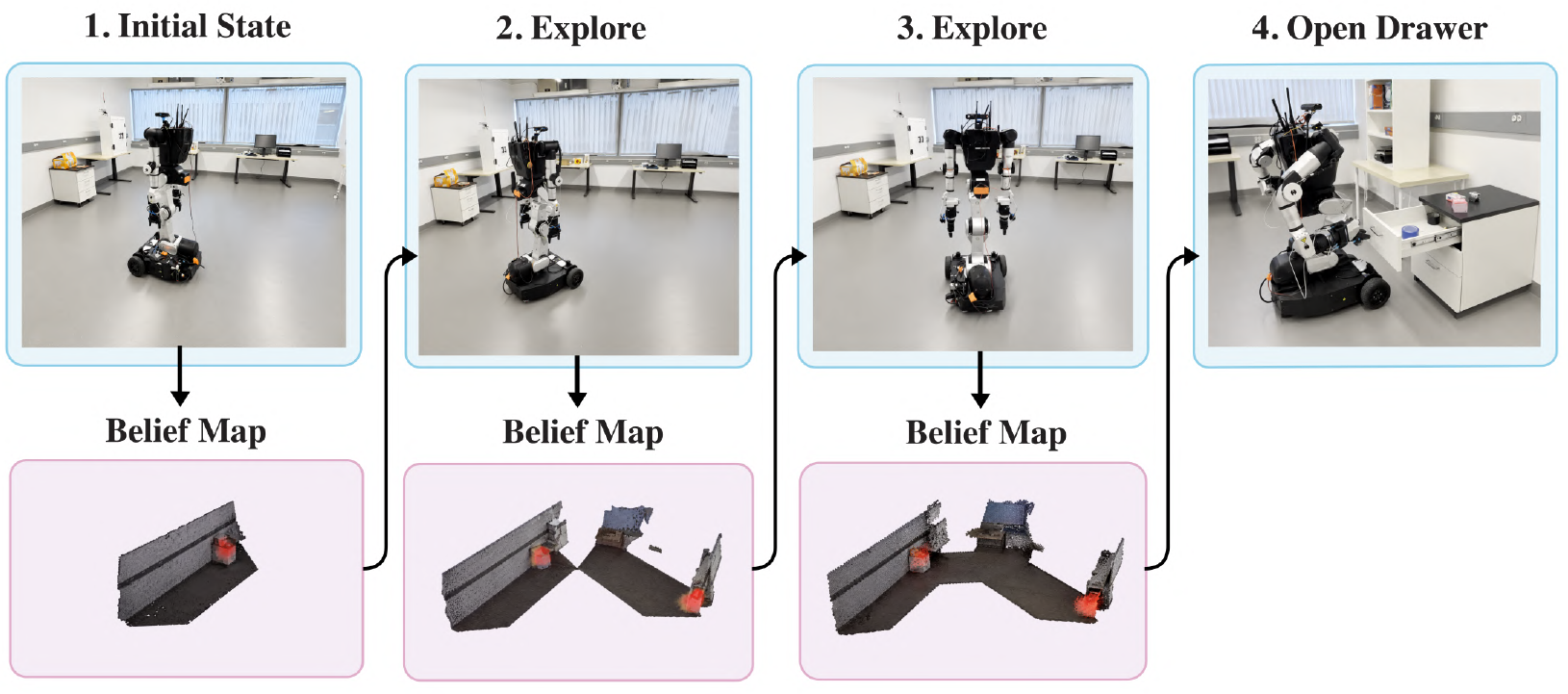}
    \vspace{-12pt}
    \caption{\textbf{Belief progression and action selection.}
Starting from the hard viewpoint, \sysname{} searches for duct
tape inside a closed drawer, executing two exploration actions
before opening it. The top row shows robot states and actions;
the belief maps below guide the selection of subsequent actions.}
\label{fig:belief_progression}
    \vspace{-12pt}
\end{figure}

\begin{figure*}[t]
    \centering
    \includegraphics[width=1\textwidth]{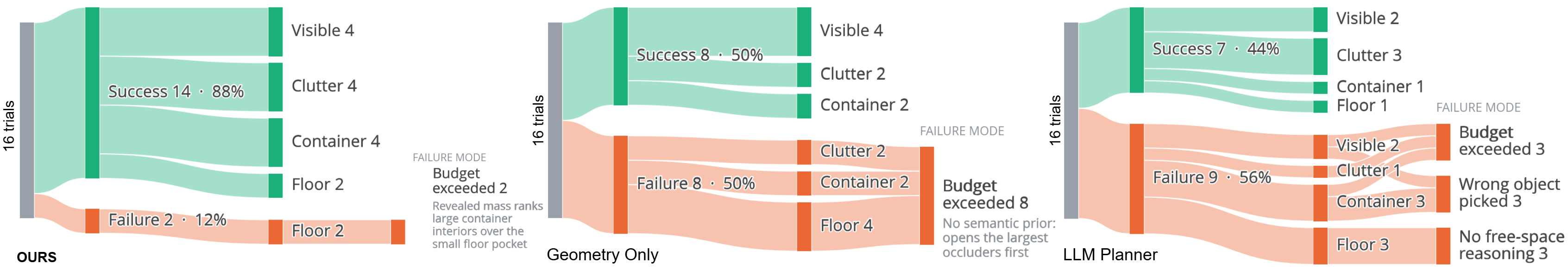}
    \vspace{-12pt}
    \caption{\textbf{Outcomes and failure modes.}
Results for \sysname{} (left), the geometry-only baseline
(middle), and the LLM planner (right), with 16 trials per method. Flows connect trial outcomes to
task settings and, for failed trials, to failure categories. Flow
widths represent trial counts; percentages are computed over the
16 trials shown for each method.}
\label{fig:failure_analysis}
    \vspace{-12pt}
\end{figure*}

\subsection{Quantitative Results}
We report two metrics: (1) \textbf{Success rate} (SR$\uparrow$), as the number of successful retrievals of the target object over total trials within the time budget; and (2) \textbf{Number of actions} (NA$\downarrow$), as the average number of actions over successful trials, to measure task efficiency. 

\begin{figure}[h]
\makeatletter
\let\c@figure\c@table\renewcommand{\thefigure}{\thetable}\renewcommand{\theHfigure}{table.\thetable}%
\renewcommand{\fnum@figure}{Table~\thetable}%
\makeatother
\centering
\renewcommand{\arraystretch}{0.85}
\small
\begin{tabular}{lcccccc}
\toprule
& \multicolumn{2}{c}{\textbf{Ours}} & \multicolumn{2}{c}{\textbf{Geometry-only}} & \multicolumn{2}{c}{\textbf{LLM-planner}} \\
\cmidrule(lr){2-3} \cmidrule(lr){4-5} \cmidrule(lr){6-7}
\textbf{Setting} & SR $\uparrow$ & NA $\downarrow$ & SR $\uparrow$ & NA $\downarrow$ & SR $\uparrow$ & NA $\downarrow$ \\
\midrule
Visible      & 4/4 & 2.5 & 4/4 & 2.5 & 2/4 & 2.0 \\
Clutter      & 4/4 & 4.0 & 2/4 & 6.0 & 3/4 & 3.0 \\
Container    & 4/4 & 3.5 & 2/4 & 4.5 & 1/4 & 5.0 \\
Floor & 2/4 & 5.0 & 0/4 & --  & 1/4 & 4.0 \\
\midrule
\textbf{Overall} & \textbf{14/16} & 3.6 & 8/16 & 3.9 & 7/16 & 3.1 \\
\bottomrule
\end{tabular}
\caption{Quantitative results with a 15-minute search budget. SR is reported as successful trials over total trials; each setting comprises 2 object categories $\times$ 2 viewpoints. NA is averaged over successful trials only.}
\label{tab:results}
\vspace{-12pt}
\end{figure}

We show that \sysname{} improves retrieval success across all the evaluated configurations, succeeding in 14/16 trials, compared with 8/16 for geometry-only and 7/16 for the LLM planner across identical settings, as shown in Table~\ref{tab:results}. 
\sysname{} succeeds in all visible, clutter, and container trials, while floor placement tasks remain challenging for all the methods. 

\sysname{} averages 3.6 actions per successful trial, while geometry-only and LLM planner average 3.9 and 3.1 actions respectively. These averages cover 14, 8, and 7 successful trials respectively. \sysname{} achieves higher success with comparable action counts among successful trials. We analyze the failures of these methods in Sec.~\ref{sec:failure_analysis}.

To see whether \sysname{}'s improvement in success rate is statistically significant, we compared paired trial outcomes, using two-sided exact McNemar test. Against geometry-only baseline, \sysname{} succeeds in six configurations where the baseline fails, with no configurations where the baseline succeeds and \sysname{} fails ($p=0.031$). Against the LLM planner, the corresponding counts are seven and zero ($p=0.016$). Both p-values fall below the significance threshold of $0.05$ providing statistical evidence that our method achieves higher SR than both baselines.

\subsection{Belief Progression and Action Selection}
We now show how \sysname{} accumulates belief over an example task of ``retrieve the duct tape'', where the tape is hidden in a storage drawer on the office side of the room, and the robot starts from the hard initial viewpoint as shown in Fig.~\ref{fig:belief_progression}. The task requires exploring an initially unseen region of the workspace and opening a storage container to reveal its contents. 

The robot starts facing the snack area. Its initial belief places belief mass on the snack-side storage drawer, alongside geometric mass on the unexplored office (not visualized, for clarity). The \aobserve{} action toward the unexplored region \texttt{explore\_at} carries the largest expected revealed belief mass $M_{\mathrm{rev}}$, so the robot takes it, revealing part of the office. The updated belief drives a second \aobserve{} action, after which an \aopen{} action on the office storage drawer, now carries the largest $M_{\mathrm{rev}}$. The robot executes the \texttt{open\_drawer} skill, revealing the duct tape, which it then retrieves. On the same task, geometry-only baseline exhaustively uncovers the largest occluded volumes, opening the cabinet and the wrong drawer, before exceeding the budget, while the LLM planner, unable to reason effectively about unseen space, interacts with the objects already in the scene graph, before committing to pick the wrong object. This sequence illustrates how \sysname{} uses the fused geometric and semantic belief update (Sec.~\ref{sec:belief_representation}) to drive action exploration and physical interaction (Sec.~\ref{sec:action_selection}) as the task progresses.

\subsection{Failure Analysis}
\label{sec:failure_analysis}
We show the failures across all three methods in Fig.~\ref{fig:failure_analysis}. The geometry-only baseline lacks semantic guidance, and spends the search budget exploring regions or interacting with containers unrelated to the target. It succeeds only when the target is clearly visible or happens to lie in the largest occluded container. 
The LLM-planner favors investigating already-observed objects in its scene-graph over exploring unseen space, and struggles under hard-viewpoint initialization. E.g., when searching for a book in the bookshelf facing the kitchen, it opens the snack drawer and the kitchen cabinet, then inspects both tables, and times out before reaching the bookshelf. Additionally, for retrieval, the planner uses the scene-graph nodes whose labels match the query, that come from an open-vocabulary detector, and verified by the VLM. From a distant view, an orange textbook can be labeled as a snack box, leading to a wrong pick. \sysname{}'s two failures are from the glue-bottle trials for the floor setting. The bottle lies in a small occluded pocket of floor beside the office drawer. This space overlaps with no object in the \obm{}, and receives no semantic evidence, and carries a small geometric prior. The fused belief isn't wrong about where the target could be, but since $M_{\mathrm{rev}}(a)$ sums belief over $\mathcal{F}(a)$, actions revealing large belief regions outrank those exposing smaller regions, supported primarily by geometric evidence.

\section{CONCLUSION}
\label{sec:conclusion}

We presented a belief-driven interactive object search system that fuses VLM-derived semantic priors with geometric evidence from occupancy raycasting into a unified 3D voxel belief field.
We demonstrated that jointly reasoning over semantics and geometry in an explicit 3D belief enables more efficient interactive object search than either modality alone.
Rather than delegating action selection to a language model over a scene graph, or exhaustively exploring the space without semantic guidance, our system grounds decisions in a voxel belief field, selecting actions toward regions most likely to contain the target by maximizing expected revealed belief mass.
Our system is deployed end-to-end on a bimanual wheeled humanoid with learned interaction skills, open-vocabulary pick-and-place, and mobile navigation, and achieves higher success-rate and more efficient search over both geometry-only and LLM-planner based baselines.

\section*{ACKNOWLEDGMENT}

We gratefully acknowledge support from NSF grant 2214177; from AFOSR grant FA9550-22-1-0249; from ONR MURI grants N00014-22-1-2740 and N00014-24-1-2603; from the MIT Siegel Family Quest for Intelligence; and from the Robotics and AI Institute.

\clearpage  %
\bibliographystyle{IEEEtran}
\bibliography{IEEEabrv,references}

\end{document}